\documentclass[runningheads]{llncs}
\usepackage[T1]{fontenc}
\usepackage{graphicx}
\usepackage{booktabs}
\usepackage[misc]{ifsym}

\usepackage{multirow}
\usepackage{amsmath}
\usepackage{graphicx}
\usepackage{subcaption}
\usepackage{amssymb}

\newcommand{\cofirst}{\textsuperscript{†}}

\begin{document}

\title{Adaptive and Explicit \texttt{<safe>}: Triggering Latent Safety Awareness in Large Reasoning Models}
\titlerunning{Triggering Latent Safety Awareness in Large Reasoning Models}

\toctitle{Adaptive and Explicit <safe>: Triggering Latent Safety Awareness in Large Reasoning Models}
\tocauthor{Ke Miao, Jiaxin Li, Hongliang Chen, Yuke Hu, Zhan Qin}


\author{Ke Miao\inst{1,2}\cofirst \and
Jiaxin Li\inst{3,4}\cofirst \and
Hongliang Chen\inst{3} \Letter \and
Yuke Hu\inst{1,2,5} \and
Zhan Qin\inst{1,2}}



\authorrunning{Ke Miao et al.}

\institute{The State Key Laboratory of Blockchain and Data Security, Zhejiang University \email{\{miaoke,yukehu,qinzhan\}@zju.edu.cn}
\and
Hangzhou HighTech Zone (Binjiang) Blockchain and Data Security Research Institute, China
\and
Li Auto Inc.
\email{\{chenhongliang,lijiaxin13\}@lixiang.com}
\and
Tsinghua University
\and
King Abdullah University Of Science And Technology
\email{yuke.hu@kaust.edu.sa}
}

\maketitle              

\let\thefootnote\relax          
\footnotetext{\textsuperscript{†} Equal contribution.}


\begin{abstract}
While Large Reasoning Models (LRMs) excel at complex tasks, they remain highly vulnerable to sophisticated jailbreaks and direct harmful queries. To address this vulnerability, prior works depend heavily on external manual data annotation for safety alignment. However, we observe that LRMs can inherently identify safety risks when being re-presented with original queries alongside their own reasoning trajectories—a capability we term Latent Safety Awareness. To leverage this safety awareness, we first employ Supervised Fine-Tuning (SFT) to explicitly induce \texttt{<safe>} tags to trigger safety analysis and guidance following the initial reasoning content for unsafe queries, while preserving standard responses for general queries to ensure adaptive triggering. Subsequently, we apply Direct Preference Optimization (DPO) to further enhance the correctness and stability of the safety analysis and guidance. Notably, responses required for both training stages are entirely generated by models being optimized. With (Safe Trigger) SFT and DPO, experimental results demonstrate significant safety enhancement. For example, the Attack Success Rate (ASR) of DeepSeek-R1-Distill-Llama-8B, on average, drops 24.65\% and 36.72\% on harmful and jailbreak benchmarks, respectively. Finally, our Safe Trigger method exerts almost no negative impact on general performance or user experience.




\end{abstract}

\section{Introduction}

Large Reasoning Models (LRMs) have recently demonstrated remarkable proficiency across both general and complex domains, acting as the major driver for new AI applications~\cite{deepseekai2025deepseekr1incentivizingreasoningcapability,openai2024openaio1card,kimiteam2025kimik15scalingreinforcement}. Despite their power, LRMs remain susceptible to generating harmful content with sophisticated jailbreaks and direct harmful queries~\cite{zhou2025hiddenriskslargereasoning,jiang2025safechain}. While safety alignment is a cornerstone of AI research, the prevailing literature focuses on standard Large Language Models (LLMs)~\cite{zhang2025stair,qi2024safetyalignmentjusttokens,zhao2025improvingllmsafetyalignment,li2025salorasafetyalignmentpreservedlowrank,liu2026aligned}, whereas specialized methodologies for LRMs are relatively underexplored. Furthermore, prior works~\cite{wang2025star,jiang2025safechain,jeung2025safepath} are heavily reliant on resource-intensive external manual annotations and leave room for improving safety robustness (as shown in our result Table~\ref{tab:performance}).

In this paper, we uncover that LRMs possess the latent safety awareness capable of identifying safety risks when re-evaluating the original query alongside their own reasoning. However, this safety awareness often remains dormant during the standard generation process. Motivated by this observation, we propose Safe Trigger (Section ~\ref{sec:safe_trigger_method}), a method designed to explicitly elicit the model’s latent safety consciousness through a structured trigger mechanism (denoted as an italicized term \textit{<safe>} in this work) utilizing \texttt{<safe>} tags. This approach strengthens the model's robustness against adversarial jailbreaks and harmful queries while preserving its general capabilities and user experience. As illustrated in Figure~\ref{fig:result}, our models—ST-S and ST-D, after Safe Trigger SFT and DPO stages, respectively—demonstrate superior safety alignment without compromising task performance. The main contributions of this work can be summarized as follows:

\begin{figure*}[t]
    \centering
    \includegraphics[width=\textwidth]{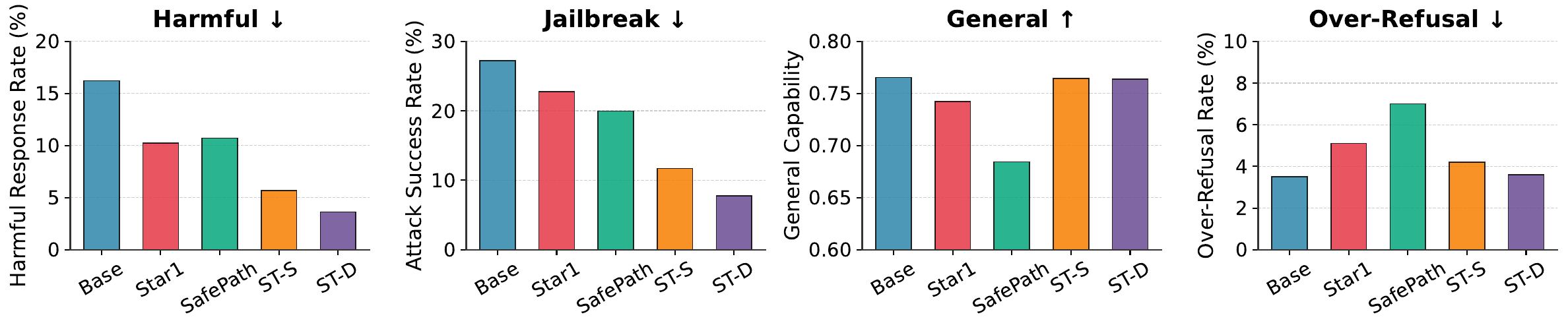}
    \caption{Performance of the model trained with our Safe Trigger approach.}
    \label{fig:result}
\end{figure*}

Our contributions are summarized as follows: \textbf{(1)} We reveal the Latent Safety Awareness phenomenon, where LRMs can identify safety issues by reviewing their own reasoning process even when an initial attack is successful. \textbf{(2)} We propose the Safe Trigger approach, utilizing Safe Trigger SFT to reliably activate a structured safety module and Safe Trigger DPO to robustly integrate safety constraints into the final output. \textbf{(3)} Extensive experiments demonstrate that our approach significantly strengthens model safety—increasing the average alignment rate of DeepSeek-R1-Distill-Llama-8B by 24.65\% on harmful benchmarks and 36.72\% on jailbreak benchmarks —while preserving general capabilities. \textbf{(4)} Our training pipeline is self-consistent, serving as its own teacher to construct signals without manual annotation or dependence on closed-source models, offering a low-cost, reproducible, and scalable solution for large-scale safety alignment.

\section{The Latent Safety Awareness of LRMs}

\textbf{Vulnerability Observations.} 
To demonstrate the limitations of current LRMs in safety alignment, we evaluate the safety of Qwen3-8B~\cite{yang2025qwen3technicalreport} and DeepSeek-R1-Distill-Llama-8B~\cite{deepseekai2025deepseekr1incentivizingreasoningcapability}, both of which serve as representative open-source LRMs. Table~\ref{tab:asr} summarizes the Attack Success Rate (ASR) of these two models across four safety evaluation datasets, where ASR reflects the proportion of queries that can cause the models to generate unsafe content.
\begin{table}[t]
\centering
\begin{minipage}{0.48\textwidth}
    \centering
    \caption{Attack Success Rate (ASR) of different LRMs across harmful and jailbreak benchmarks when original queries are entered directly.}
    \label{tab:asr}
    \resizebox{\linewidth}{!}{%
    \begin{tabular}{l|ccc|c}
    \toprule
    \multirow{2}{*}{\textbf{Model}} & \multicolumn{3}{c|}{\textbf{Harmful}} & \textbf{Jailbreak} \\ \cline{2-5}
     & Advbench & HexPHI & XsTest & WildJailbreak \\ \midrule
    Qwen3-8B & 1.92\% & 13.33\% & 0.50\% & 42.80\% \\
    DeepSeek-8B & 35.19\% & 44.00\% & 20.00\% & 49.90\% \\ \bottomrule
    \end{tabular}%
    }
\end{minipage}
\hfill 
\begin{minipage}{0.48\textwidth}
    \centering
    \caption{Risk Identification Success Rate (RISR) of different LRMs when re-evaluating original queries and their corresponding reasoning trajectories.}
    \label{tab:risr}
    \resizebox{\linewidth}{!}{%
    \begin{tabular}{l|ccc|c}
    \toprule
    \multirow{2}{*}{\textbf{Model}} & \multicolumn{3}{c|}{\textbf{Harmful}} & \textbf{Jailbreak} \\ \cline{2-5}
     & Advbench & HexPHI & XsTest & WildJailbreak \\ \midrule
    Qwen3-8B & 60.00\% & 75.00\% & 100.00\% & 58.29\% \\
    DeepSeek-8B & 81.97\% & 69.70\% & 87.50\% & 44.79\% \\ \bottomrule
    \end{tabular}%
    }
\end{minipage}
\end{table}


LRMs exhibit pronounced vulnerabilities to generating harmful outputs, highlighting a critical deficiency in their safety alignment. The results in Table~\ref{tab:asr} show that Qwen3-8B already exhibits a certain probability of generating unsafe content in the context of harmful queries, with the highest attack success rate being 13.33\% on the HexPHI dataset~\cite{qi2023finetuningalignedlanguagemodels}. 
In the context of jailbreak queries, the probability increases sharply, reaching 42.80\% for the WildJailbreak dataset~\cite{jiang2024wildteamingscaleinthewildjailbreaks}. DeepSeek-R1-Distill-Llama-8B suffers from a high ASR on all four benchmarks. 



\textbf{Latent Safety Awareness.} Despite the vulnerabilities identified above, our further experiments uncovered a compelling phenomenon: LRMs possess a latent capacity for safety risk identification that often remains dormant during standard generation. Specifically, we conducted a diagnostic study on the subset of queries that successfully bypassed the model's safety guard to elicit harmful content. We re-presented the adversarial queries alongside the model's self-generated reasoning trajectories to the model itself for a safety audit (quantified via safety risk identification), tasking it to assess the adequacy of its own safety considerations regarding the input and reasoning content. As shown in Table~\ref{tab:risr}, we report the Risk Identification Success Rate (RISR), which quantifies the model's ability to post-hoc identify safety risks. The fact that RISR ranges from 44.79\% to 100\% (consistently surpassing the ASR) suggests that LRMs possess a significant latent capacity to recognize safety risks within both the original queries and their own reasoning trajectories, which offers a promising foundation for safety alignment in our work.

The above results demonstrate that when LRMs are presented with both the initial adversarial queries and their own reasoning paths, their capacity to identify core safety violations and internal vulnerabilities increases significantly. Under these conditions, the models can proactively recommend refusing to continue generating potentially harmful answers. We define this capability as the \textbf{Latent Safety Awareness} of LRMs. However, in the default reasoning process of existing LRMs, there is no safety reminder mechanism between the end of reasoning and the generation of the final response for risky queries. Even though the model has Latent Safety Awareness, it is often not effectively activated in the standard process.
Based on the aforementioned findings, we propose the \textbf{Safe Trigger} approach. It systematically activates and enhances the model's Latent Safety Awareness and significantly improves the model's safety capabilities without compromising its general performance. 

\section{Our Safe Trigger Method}\label{sec:safe_trigger_method}

\begin{figure*}[t]
    \centering
    \includegraphics[width=\textwidth]{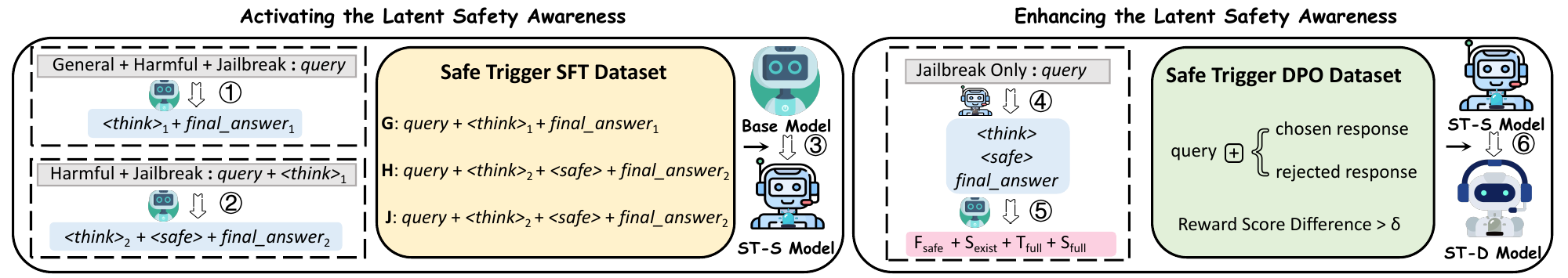}
    \caption{Overview of our Safe Trigger method. The two-stage process activates and enhances the model's Latent Safety Awareness. (a) Safe Trigger SFT: The model is trained on general, harmful, and jailbreak queries to adaptively elicit a structured trigger mechanism (\textit{<safe>}), which provides both safety risk analysis and response generation guidance. (b) Safe Trigger DPO: The model refines its safety discernment by optimizing on self-generated preference pairs. These pairs are derived from jailbreak queries and ranked using our reward function (Eq.~\ref{eq:reward_final}).}
    \label{fig:flow}
\end{figure*}

To explicitly activate and effectively leverage Latent Safety Awareness, we introduce Safe Trigger SFT to make LRMs explicitly activate the structured trigger mechanism (denoted as an italicized term \textit{<safe>}), considering safety risks and guiding answer generation within \texttt{<safe>} and \texttt{</safe>} tags for unsafe queries. To avoid possible ambiguity, we denote the structured trigger mechanism as \textit{<safe>} (in italics), while tags are represented as \texttt{<safe>} and \texttt{</safe>}.
Besides, we propose Safe Trigger DPO to further enhance the guiding effectiveness and correctness of the structured trigger mechanism on the final output, reinforcing the model’s ability to produce more instructive safety content.
The overall training pipeline of the Safe Trigger approach is illustrated in Figure~\ref{fig:flow}.


\subsection{Safe Trigger SFT: Activate the Latent Safety Awareness}

For risky or uncertain queries, the structured trigger mechanism (\textit{<safe>}) is inserted between the end of the reasoning process and the final response in the form of \texttt{<safe>}~\texttt{</safe>}, serving as an intermediate step prior to the final answer.
This mechanism consists of the following three components:
\begin{itemize}
    \item \textbf{Core Issue Safety Analysis}: The model re-examines the core issue based on the query and its own reasoning process, and determines whether the issue poses potential safety risks.
    
    \item \textbf{Reasoning Process Safety Analysis}: If potential risks are identified, the model further reviews whether such risks have been adequately considered during the reasoning stage.
    
    \item \textbf{Final Answer Guidance Content}: Based on the above analyses, the model generates guidance for the final answer, encouraging the production of more appropriate response.
\end{itemize}
We constructed a training dataset consisting of 30k instances, evenly partitioned into three categories: general, harmful, and jailbreak queries (10k per category). These instances were sourced from the UltraFeedback~\cite{cui2024ultrafeedback}, PKU-SafeRLHF~\cite{ji2024pku}, and WildJailbreak~\cite{jiang2024wildteamingscaleinthewildjailbreaks} datasets. 



We employ a two-stage generation strategy to construct training samples.
In the first stage, we directly input the $query$ into the model to be optimized to obtain an initial reasoning result, which includes the reasoning process \textit{<think>}$_1$ and the corresponding $final\ answer_1$. 
In the second stage, we input the risky $query$ including harmful and jailbreak queries along with \textit{<think>}$_1$ into the same model, and append a safety analysis prompt to guide the model in performing a safety assessment based on its reasoning process and the risky $query$. 
The output consists of a regenerated reasoning process \textit{<think>}$_2$, the structured trigger mechanism \textit{<safe>}, and $final\ answer_2$. 


We construct training samples according to the following rules. For general samples, the data follows the structure $\{\textit{query} + \textit{<think>}_1 + \textit{final\_answer}_1\}$. For harmful and jailbreak samples, the data follows the structure $\{\textit{query} + \textit{<think>}_1 + \textit{<safe>} + \textit{final\_answer}_2\}$. The combination of general, harmful, and jailbreak samples in the dataset, along with the omission of the structured trigger mechanism (\textit{<safe>}) in general samples, enables the model to adaptively activate the structured trigger mechanism (\textit{<safe>}). The harmful and jailbreak samples teach the model what analysis should be included within the structured trigger mechanism. After training with Safe Trigger SFT, we obtain the Safe Trigger SFT Model (ST-S).


\subsection{Safe Trigger DPO: Enhance the Latent Safety Awareness}
The ST-S model's safety capabilities have been significantly improved compared with the base model. 
However, due to the inherent limitations of Latent Safety Awareness, there are still cases where, even after activating the structured trigger mechanism, the model fails to effectively guide the generation of a safe final answer. 
To address this issue, we further introduce the Safe Trigger DPO.
We use 20k jailbreak queries for preference optimization, all selected from the adversarial harmful subset of the WildJailbreak dataset and disjoint from the jailbreak queries used in the SFT stage. 

\textbf{Binary Variables and Target Ranking.} For each structured model response, we define four binary indicators: 
\begin{equation}
F_{\text{safe}},\ S_{\text{exist}},\ T_{\text{full}},\ S_{\text{full}} \in \{0,1\}
\end{equation}
where $F_{\text{safe}}$ indicates whether the \emph{final answer} is safe, $S_{\text{exist}}$ indicates whether the response contains a complete structured trigger mechanism (\textit{<safe>}), $T_{\text{full}}$ indicates whether the reasoning trace sufficiently considers safety, and $S_{\text{full}}$ indicates whether the analysis in \textit{<safe>} is comprehensive.
Our reward function enforces the following preference hierarchy among four response types:
\begin{align}
\textbf{A: } &\ \text{Safe final answer with \textit{<safe>}} \nonumber \ 
\hfill (F_{\text{safe}}=1,\ S_{\text{exist}}=1) \nonumber\\
\textbf{B: } &\ \text{Safe final answer without \textit{<safe>}} \nonumber\ 
\hfill (F_{\text{safe}}=1,\ S_{\text{exist}}=0) \nonumber\\
\textbf{C: } &\ \text{Unsafe final answer with \textit{<safe>}} \nonumber\ 
\hfill (F_{\text{safe}}=0,\ S_{\text{exist}}=1) \nonumber\\
\textbf{D: } &\ \text{Unsafe final answer without \textit{<safe>}} \nonumber\ 
\hfill (F_{\text{safe}}=0,\ S_{\text{exist}}=0) \nonumber
\end{align}

In particular, we want:
\begin{equation}
\textbf{A} \succ \textbf{B} \succ \textbf{C} \succ \textbf{D}
\label{eq:target_ranking}
\end{equation}

\textbf{Piecewise Construction.} We first define a reward function in a piecewise manner by conditioning on $(F_{\text{safe}}, S_{\text{exist}})$.

Case A: Safe final answer with \textit{<safe>} ($F_{\text{safe}}=1, S_{\text{exist}}=1$).
When the final answer is safe and \textit{<safe>} is activated, we additionally reward the \emph{quality} of the safety-related reasoning and analysis. This is captured by the sum $(T_{\text{full}} + S_{\text{full}})$, and we add a base offset $w_a$ to ensure this case dominates the other cases:
\begin{equation}
R_A \triangleq T_{\text{full}} + S_{\text{full}} + w_a
\label{eq:RA}
\end{equation}

Case B: Safe final answer without \textit{<safe>} ($F_{\text{safe}}=1, S_{\text{exist}}=0$).
If the final answer is safe but \textit{<safe>} is not activated, we assign a constant reward $w_b$:
\begin{equation}
R_B \triangleq w_b
\label{eq:RB}
\end{equation}
This still rewards safety, but it is intentionally lower than Case A so that, among safe answers, activating \textit{<safe>} is preferred.

Case C: Unsafe final answer with \textit{<safe>} ($F_{\text{safe}}=0, S_{\text{exist}}=1$).
If \textit{<safe>} is activated but the final answer remains unsafe, we assign a smaller reward $w_c$:
\begin{equation}
R_C \triangleq w_c
\label{eq:RC}
\end{equation}
This makes unsafe outputs strictly worse than safe outputs, even when \textit{<safe>} exists.

Case D: Unsafe final answer without \textit{<safe>} ($F_{\text{safe}}=0, S_{\text{exist}}=0$).
This is the completely unacceptable failure mode (unsafe and no structured trigger mechanism). We collapse the reward to zero:
\begin{equation}
R_D \triangleq 0
\label{eq:RD}
\end{equation}

\textbf{Unified Closed-Form Expression.} The above piecewise definition can be written as a single closed-form expression using indicator products.
Let $\mathbb{I}[\cdot]$ be the indicator function. Since $F_{\text{safe}}, S_{\text{exist}}\in\{0,1\}$, we can use products directly as indicators:
\begin{equation}
\mathbb{I}[F_{\text{safe}}=1] = F_{\text{safe}},\ \mathbb{I}[S_{\text{exist}}=1] = S_{\text{exist}}
\end{equation}
\begin{equation}
\mathbb{I}[S_{\text{exist}}=0] = (1-S_{\text{exist}}),\ \mathbb{I}[F_{\text{safe}}=0] = (1-F_{\text{safe}})
\end{equation}
Combining Eqs.~\eqref{eq:RA}--\eqref{eq:RD}, the reward becomes:
\begin{equation}
\begin{aligned}
R
=\ &\mathbb{I}[F_{\text{safe}}=1,S_{\text{exist}}=1]\cdot (T_{\text{full}} + S_{\text{full}} + w_a)
+ \mathbb{I}[F_{\text{safe}}=1,S_{\text{exist}}=0]\cdot w_b \\
&+ \mathbb{I}[F_{\text{safe}}=0,S_{\text{exist}}=1]\cdot w_c
+ \mathbb{I}[F_{\text{safe}}=0,S_{\text{exist}}=0]\cdot 0
\end{aligned}
\label{eq:reward_indicator}
\end{equation}
which simplifies to:
\begin{equation}
\begin{aligned}
R =\ &F_{\text{safe}} \cdot  S_{\text{exist}} \cdot (T_{\text{full}} + S_{\text{full}} + w_a)
    + F_{\text{safe}}(1 - S_{\text{exist}}) \cdot w_b \\
    &+ (1 - F_{\text{safe}}) \cdot S_{\text{exist}} \cdot w_c
\end{aligned}
\label{eq:reward_final}
\end{equation}

\textbf{Constraints and Ranking Guarantees.} We impose $\boldsymbol{w_a > w_b > w_c}$ to reflect the intended preference structure in Eq.~\eqref{eq:target_ranking}. We show this more explicitly below.

A is better than B (safe with activating vs.\ safe without activating).
For Case A, $R_A = T_{\text{full}} + S_{\text{full}} + w_a \ge w_a$ since $T_{\text{full}}, S_{\text{full}}\ge 0$.
For Case B, $R_B = w_b$.
Thus, $R_A > R_B$ is guaranteed by:
\begin{equation}
w_a > w_b
\end{equation}

A is better than C (safe with activating vs.\ unsafe with activating).
Case A has $R_A \ge w_a$, while Case C has $R_C = w_c$.
Thus, $R_A > R_C$ is guaranteed by:
\begin{equation}
w_a > w_c
\end{equation}

B is better than C (safe without activating vs.\ unsafe with activating).
Case B has $R_B = w_b$, while Case C has $R_C = w_c$.
Thus, $R_B > R_C$ is guaranteed by:
\begin{equation}
w_b > w_c
\end{equation}

D collapses to zero (unsafe without activating).
When $F_{\text{safe}}=0$ and $S_{\text{exist}}=0$, each term in Eq.~\eqref{eq:reward_final} becomes zero:
\begin{equation}
R = 0 \cdot 0 \cdot (\cdot) + 0\cdot 1 \cdot w_b + 1\cdot 0 \cdot w_c = 0
\end{equation}
Therefore, Case D always receives $R_D=0$, matching the intended ``completely unacceptable'' behavior.

Given $F_{\text{safe}}=1$ and $S_{\text{exist}}=1$, the reward additionally increases with $T_{\text{full}}$ and $S_{\text{full}}$. This encourages not only activating but also producing safety-aware reasoning traces and a comprehensive safety analysis, thereby improving the guiding power and stability of \textit{<safe>} during preference optimization.

\textbf{Detailed Training Procedure.}
For the 20k jailbreak queries, we first perform high temperature sampling to generate diverse responses. For each query $q_i$, we sample 4 distinct structured outputs $\{r_{i,1}, r_{i,2}, r_{i,3}, r_{i,4}\}$, each consisting of a reasoning trace, the structured trigger mechanism (if triggered), and a final answer.
Each response $r_{i,j}$ is independently scored using the reward function $R(r_{i,j})$. 
We select queries where the sampled responses exhibit significant variance in reward scores. Formally, we retain only those queries $q_i$ for which:
\begin{equation}
\max_j R(r_{i,j}) - \min_j R(r_{i,j}) \geq \delta \quad 
\end{equation}
This filtering ensures that each retained query provides both a high-quality (positive) and a low-quality (negative) response, denoted as $(r_i^{+}, r_i^{-})$.
These preference pairs are used to train the model. The training objective maximizes the preference for the better response over the worse one, relative to a reference policy $\pi_{\text{ref}}$, and is defined as:

\begin{equation}
\mathcal{L}_{\text{DPO}} = - \log \sigma \left( \beta \cdot \left[ \log \frac{\pi_{\theta}(r_i^{+} \mid q_i)}{\pi_{\text{ref}}(r_i^{+} \mid q_i)} - \log \frac{\pi_{\theta}(r_i^{-} \mid q_i)}{\pi_{\text{ref}}(r_i^{-} \mid q_i)} \right] \right)
\end{equation}


where $\pi_{\theta}$ is the current policy, $\pi_{\text{ref}}$ is the reference policy, and $\beta$ controls the sharpness of preference. After training with Safe Trigger DPO, we obtain the Safe Trigger DPO Model (ST-D).

\section{Experimental Results}

\subsection{Experimental Settings}
We conduct experiments on four representative LRMs in the main comparison, including Qwen3-8B, Qwen3-32B, DeepSeek-R1-Distill-Llama-8B, and DeepSeek-R1-Distill-Llama-70B~\cite{yang2025qwen3technicalreport,deepseekai2025deepseekr1incentivizingreasoningcapability}. For additional experiments and analyses, we performed them on our default backbone model, DeepSeek-R1-Distill-Llama-8B.

\begin{table*}[t]
\centering
\caption{Performance comparison across different models and methods. ↓ indicates lower is better, ↑ indicates higher is better. 
\textbf{Bold} indicates the best result, and \underline{underline} indicates the second best. 
These notations are consistently applied in all subsequent tables.}
\label{tab:performance}
\resizebox{\textwidth}{!}{%
\begin{tabular}{l|ccc|ccc|ccc|c}
\toprule
\multirow{2}{*}{\textbf{Method}} & \multicolumn{3}{c|}{\textbf{Harmful ↓}} & \multicolumn{3}{c|}{\textbf{Jailbreak ↓}} & \multicolumn{3}{c|}{\textbf{General ↑}} & \textbf{Over-Refusal ↓} \\ \cline{2-11}
 & AdvB. & HexPHI & XsTest & WildJail. & MSJ & PAP & ARC & Drop & Wino & XsTest-S \\ \hline
\multicolumn{11}{c}{Qwen3-8B} \\ \hline
Base & 1.92\% & 13.33\% & \underline{0.50\%} & 42.80\% & 8.00\% & 12.00\% & \underline{0.9267} & \textbf{0.6623} & \textbf{0.7822} & 3.20\% \\
Star1~\cite{wang2025star} & \textbf{0.19\%} & 8.67\% & \underline{0.50\%} & 37.40\% & 10.00\% & 12.00\% & \textbf{0.9278} & 0.6364 & 0.6551 & \underline{2.40\%} \\
SafePath~\cite{jeung2025safepath} & \underline{0.38\%} & 7.33\% & \textbf{0.00\%} & 36.95\% & 18.00\% & 12.00\% & 0.9264 & 0.6226 & 0.6543 & 3.20\% \\
ST-S & 0.58\% & \underline{6.67\%} & \textbf{0.00\%} & \underline{29.50\%} & \textbf{0.00\%} & \underline{10.00\%} & 0.9250 & \underline{0.6580} & \underline{0.7782} & 4.80\% \\
ST-D & \underline{0.38\%} & \textbf{5.67\%} & \textbf{0.00\%} & \textbf{23.20\%} & \underline{2.00\%} & \textbf{4.00\%} & 0.9242 & 0.6567 & 0.7774 & \textbf{2.00\%} \\ \hline
\multicolumn{11}{c}{DeepSeek-R1-Distill-Llama-8B} \\ \hline
Base & 35.19\% & 44.00\% & 20.00\% & 49.90\% & 68.00\% & 20.00\% & \textbf{0.8810} & 0.4677 & 0.6638 & 4.00\% \\
Star1~\cite{wang2025star} & 14.04\% & 20.67\% & 9.50\% & 39.15\% & 54.00\% & 14.00\% & \underline{0.8796} & \underline{0.4837} & 0.5359 & 7.20\% \\
SafePath~\cite{jeung2025safepath} & 14.62\% & 24.67\% & 8.50\% & \textbf{16.65\%} & 58.00\% & \textbf{2.00\%} & 0.8698 & 0.4561 & 0.5130 & 12.00\% \\
ST-S & \underline{9.62\%} & \underline{20.00\%} & \underline{8.00\%} & 29.95\% & \underline{14.00\%} & \underline{4.00\%} & 0.8749 & \textbf{0.4876} & \textbf{0.6835} & \underline{2.80\%} \\
ST-D & \textbf{4.42\%} & \textbf{13.33\%} & \textbf{7.50\%} & \underline{17.75\%} & \textbf{8.00\%} & \textbf{2.00\%} & 0.8791 & 0.4620 & \underline{0.6732} & \textbf{2.40\%} \\ \hline
\multicolumn{11}{c}{Qwen3-32B} \\ \hline
Base & 3.85\% & 12.33\% & \textbf{0.00\%} & 35.40\% & 10.00\% & 12.00\% & 0.9459 & 0.6272 & \textbf{0.7648} & \textbf{1.60\%} \\
Star1~\cite{wang2025star} & 6.73\% & \underline{9.00\%} & \textbf{0.00\%} & 30.85\% & 12.00\% & \underline{10.00\%} & 0.9434 & \underline{0.6402} & 0.7403 & 3.20\% \\
SafePath~\cite{jeung2025safepath} & 4.81\% & 13.33\% & \textbf{0.00\%} & \textbf{24.80\%} & 10.00\% & \underline{10.00\%} & 0.9459 & \textbf{0.6513} & 0.7245 & \underline{2.80\%} \\
ST-S & \textbf{1.35\%} & 10.00\% & \textbf{0.00\%} & 27.15\% & \underline{4.00\%} & \underline{10.00\%} & \underline{0.9461} & 0.6081 & \underline{0.7514} & 3.20\% \\
ST-D & \underline{1.73\%} & \textbf{8.33\%} & \textbf{0.00\%} & \underline{27.10\%} & \textbf{2.00\%} & \textbf{4.00\%} & \textbf{0.9475} & 0.6081 & \textbf{0.7648} & 3.20\% \\ \hline
\multicolumn{11}{c}{DeepSeek-R1-Distill-Llama-70B} \\ \hline
Base & 26.15\% & 23.33\% & 14.00\% & 32.45\% & 18.00\% & 18.00\% & 0.9416 & 0.6721 & 0.8493 & \textbf{5.20\%} \\
Star1~\cite{wang2025star} & 18.65\% & 19.67\% & 15.50\% & 29.60\% & \underline{10.00\%} & 14.00\% & \textbf{0.9433} & \textbf{0.6733} & 0.8461 & 7.60\% \\
SafePath~\cite{jeung2025safepath} & 20.96\% & 20.00\% & 14.00\% & 28.95\% & \underline{10.00\%} & \underline{12.00\%} & 0.9411 & 0.6665 & 0.8453 & 10.00\% \\
ST-S & \underline{1.35\%} & \underline{7.33\%} & \underline{3.50\%} & \underline{9.95\%} & \textbf{0.00\%} & \textbf{2.00\%} & 0.9411 & 0.6689 & \underline{0.8516} & \underline{6.00\%} \\
ST-D & \textbf{0.58\%} & \textbf{1.67\%} & \textbf{0.00\%} & \textbf{1.45\%} & \textbf{0.00\%} & \textbf{2.00\%} & \underline{0.9417} & \underline{0.6730} & \textbf{0.8564} & 6.80\% \\ \bottomrule
\end{tabular}%
}
\end{table*}

\textbf{Baselines.} We compare the proposed Safe Trigger approach with two representative safety alignment methods for LRMs.
Star1~\cite{wang2025star} is a data-centric alignment method that improves LRM safety by fine-tuning models on a carefully curated, human-audited, small-scale dataset consisting of 1,000 high-quality safety samples.
SafePath~\cite{jeung2025safepath} is an alignment approach specifically designed for LRMs. It fine-tunes the model to emit a short safety prefix at the beginning of the reasoning process when encountering harmful prompts, while leaving the remainder of the reasoning trajectory unconstrained. SafePath is designed to reduce harmful outputs with minimal interference to reasoning depth and is regarded as the current state-of-the-art method.

\textbf{Evaluation metrics.} Four types of tasks were designed to comprehensively evaluate the methods. \textbf{(1)} Harmful task evaluates the model’s robustness against direct harmful or illegal requests. We adopt several widely used harmful benchmarks, including AdvBench~\cite{chen2022adversarialperturbationsimperceptiblerethink}, HexPHI~\cite{qi2023finetuningalignedlanguagemodels}, and XsTest (Unsafe subset)~\cite{röttger2024xstesttestsuiteidentifying}, which together cover a broad range of explicit safety risks. \textbf{(2)} Jailbreak task assesses the model’s robustness under adversarial manipulation. Specifically, we use the test set of WildJailbreak~\cite{jiang2024wildteamingscaleinthewildjailbreaks} and further evaluate the model under two strong jailbreak attack methods, MSJ~\cite{anil2024manyshot} and PAP~\cite{zeng2024johnnypersuadellmsjailbreak}. 
Following prior work~\cite{li2025salora,jiang-etal-2025-safechain,xue2025loraneedsafetyalignment}, we use LlamaGuard-3-8B as an automatic safety discriminator to determine whether the model outputs unsafe content. LlamaGuard-3-8B is a widely adopted open-source safety classifier that provides consistent, scalable, and reproducible safety judgments across different models and alignment methods. The research~\cite{li2025reasoningshieldsafetydetectionreasoning} also experimentally demonstrated that LlamaGuard-3-8B achieves higher accuracy in evaluating the final answers of LRMs. \textbf{(3)} General task is used to verify that safety alignment does not significantly degrade the model’s general performance. We evaluate the model on a set of commonly used general benchmarks, including ARC (Mean of Easy and Challenge)~\cite{allenai:arc}, DROP~\cite{dua2019dropreadingcomprehensionbenchmark}, and Wino~\cite{sakaguchi2019winogrande}, covering reasoning, reading comprehension, and common-sense understanding.
We adopt the EvalScope~\cite{evalscope_2024} evaluation framework to obtain standardized and comparable performance scores. \textbf{(4)} Over-Refusal task measures whether the model incorrectly refuses to answer problems that seem harmful but are actually harmless after safety alignment. We use the XsTest (Safe subset)~\cite{röttger2024xstesttestsuiteidentifying} benchmark to evaluate over-refusal behavior. 
We employ GPT-4o~\cite{openai2024gpt4technicalreport} as an external judge to determine whether the model unnecessarily refuses to answer safe queries. 

\textbf{Implementation.} All experiments were conducted on a single compute node equipped with 8 × NVIDIA L20X GPUs, each with 144 GB of memory. The node is powered by two Intel Xeon Platinum 8558 processors, providing 192 CPU cores and a total of 1 TB of system memory. Experiments were executed on multiple GPUs using the Hugging Face Transformer Reinforcement Learning 0.23.0 framework with LoRA-based parameter-efficient fine-tuning, and LoRA was implemented with a rank of 64, an $\alpha$ value of 16, and a dropout rate of 0.05 by default.  

\subsection{Main Results}

Table~\ref{tab:performance} reports detailed performance for each method across four LRMs and all evaluated benchmarks. Across all four LRMs, Safe Trigger consistently improves safety alignment in both harmful and jailbreak scenarios while preserving general capability. 

On harmful benchmarks, both ST-S and ST-D reduce harmful response rates relative to Base, Star1, and SafePath for most model-benchmark combinations. Safe Trigger substantially lowers the harmful response rate across multiple harmful datasets, indicating that explicitly inserting structured safety analysis between reasoning and final answer can effectively mitigate unsafe completion. On jailbreak evaluations, Safe Trigger also shows clear advantages: ST-S already provides strong reductions in attack success rate, and ST-D further improves robustness, suggesting that preference optimization strengthens the stability and guiding power of the triggered safety analysis under adversarial prompts. Meanwhile, Safe Trigger preserves general capability, remaining comparable to the Base model and avoiding the larger degradation observed in some baselines. Safe Trigger maintains a relatively low over-refusal rate on XsTest-Safe, indicating that the method does not rely on overly aggressive refusal to obtain safety gains.

To provide a more intuitive comparison, we further summarize the results by averaging performance across all evaluated models and benchmarks within each task category. The aggregated results are reported in Table~\ref{tab:performance_comparison}, which highlights the overall safety--utility trade-offs of different methods. The visualization of this table is shown in Figure~\ref{fig:result}. Overall, Safe Trigger achieves the best balance between safety and utility among all compared methods. 
\begin{table}[t]
\centering
\caption{Overall average performance comparison across different safety alignment methods. Results are computed by taking the mean over all evaluated models and all benchmarks within each category. }
\label{tab:performance_comparison}
\resizebox{0.7\columnwidth}{!}{%
\begin{tabular}{l|cccc}
\toprule
\textbf{Method} & \textbf{Harmful ↓} & \textbf{Jailbreak ↓} & \textbf{General ↑} & \textbf{Over-Refusal ↓} \\ \midrule
Base      & 16.22\% & 27.21\% & \textbf{0.7654} & \textbf{3.50\%} \\
Star1~\cite{wang2025star}     & 10.26\% & 22.75\% & 0.7421 & 5.10\% \\
SafePath~\cite{jeung2025safepath}  & 10.72\% & 19.95\% & 0.6843 & 7.00\% \\
ST-S      & \underline{5.70\%}  & \underline{11.71\%} & \underline{0.7645} & 4.20\% \\
ST-D      & \textbf{3.63\%}  & \textbf{7.79\%}  & 0.7637 & \underline{3.60\%} \\
\bottomrule
\end{tabular}%
}
\end{table}
In terms of harmful and jailbreak metrics, both ST-S and ST-D substantially outperform the baseline methods, with ST-D achieving the lowest average harmful response rate and jailbreak attack success rate. 
For general capability, Safe Trigger remains comparable to the Base model. Regarding the Over-Refusal benchmark, ST-D maintains a low refusal rate close to the Base model, suggesting that the additional preference optimization stage helps integrate safety constraints into the final answer without relying on excessive refusals.

\subsection{Detailed Analysis}

\subsubsection{Alignment Depth}

\begin{figure}[t]
    \centering
    \begin{subfigure}[b]{0.48\textwidth}
        \centering
        \includegraphics[width=\linewidth]{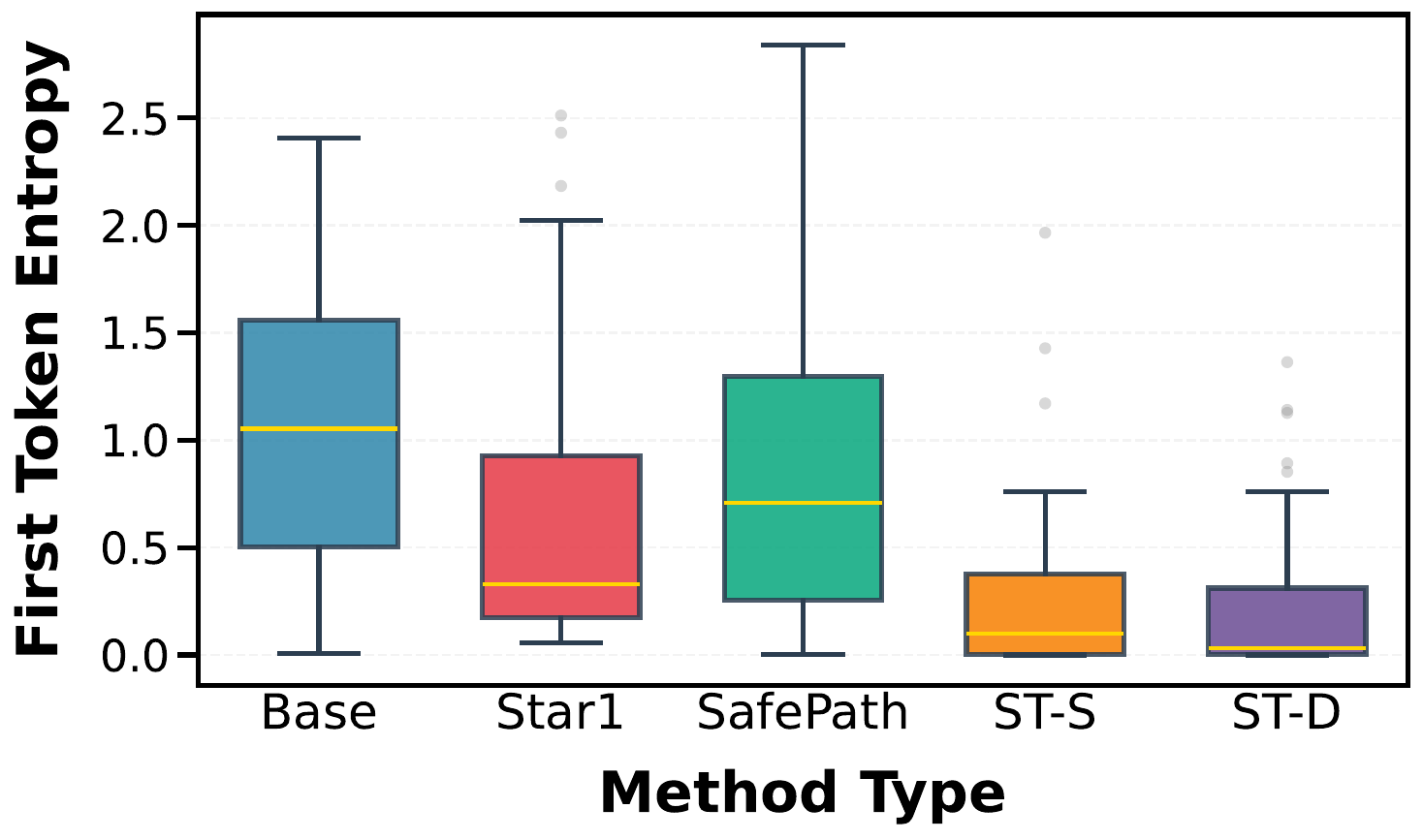}
        \caption{Entropy Distribution.}
        \label{fig:entropy_distribution}
    \end{subfigure}
    \hfill
    \begin{subfigure}[b]{0.48\textwidth}
        \centering
        \includegraphics[width=\linewidth]{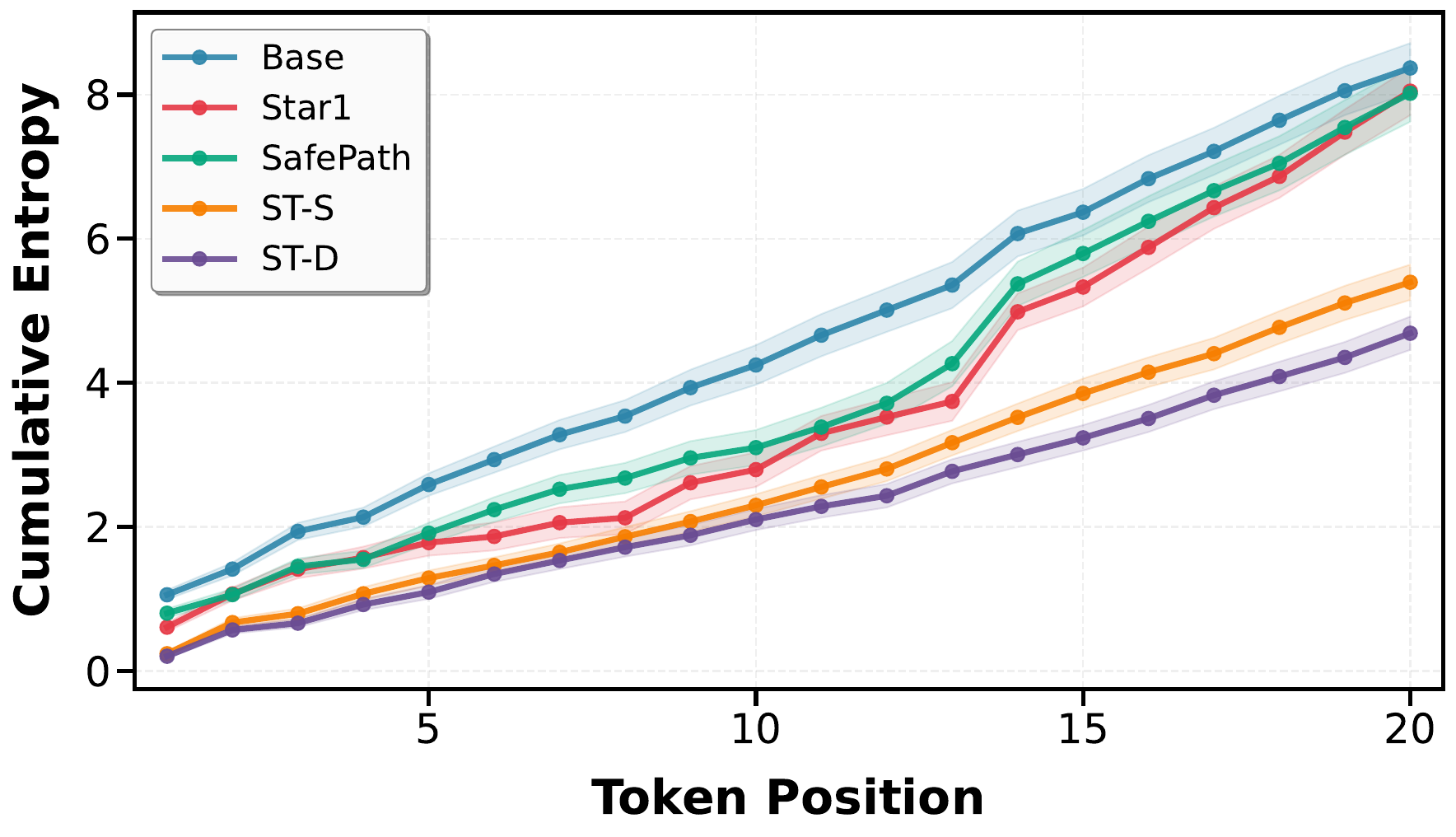}
        \caption{Cumulative Entropy.}
        \label{fig:cumulative_entropy}
    \end{subfigure}
    
    \caption{Entropy distribution and cumulative trends of the initial tokens.}
    \label{fig:alignment_depth}
\end{figure}

Figure~\ref{fig:alignment_depth} analyzes the model behavior during the early stage of final answer decoding by examining the first-token entropy distribution and the cumulative entropy across token positions~\cite{qi2024safetyalignmentjusttokens}. The data in Figure~\ref{fig:alignment_depth} are collected from 100 conditionally sampled queries from AdvBench, where we only retain queries on which all five methods (Base, Star1, SafePath, ST-S, ST-D) produce responses that are judged as safe. By controlling safety outcomes in this way, the goal is to compare which method behaves better when they all successfully produce safe answers.

Under this controlled setting, Figure~\ref{fig:entropy_distribution} shows that the base model exhibits significantly higher first-token entropy with a broad distribution, indicating less deterministic behavior at the very beginning of final answer generation. Star1 and SafePath demonstrate moderate entropy reduction compared to Base, but still present relatively wider distributions with more high-entropy outliers, suggesting inconsistent early decoding patterns. ST-S and ST-D achieve lower and more concentrated first-token entropy distributions, with ST-D showing the tightest concentration near zero entropy. This indicates that Safe Trigger exhibits highly confident and consistent first-step decisions when initiating safe responses.

Figure~\ref{fig:cumulative_entropy} further reveals the stability of safety alignment in the early generation process. The base model accumulates entropy most rapidly, reflecting sustained uncertainty across multiple decoding steps. Star1 and SafePath show moderate cumulative entropy growth, while ST-S demonstrates notably slower accumulation. ST-D maintains the lowest cumulative entropy trajectory, indicating the most stable and deterministic generation path. The diverging cumulative entropy curves suggest that Safe Trigger not only improves initial token confidence but also maintains consistently lower uncertainty in the early generation phase.

\subsubsection{Activation Probability}

\begin{figure}[t]
    \centering
    \includegraphics[width=\columnwidth]{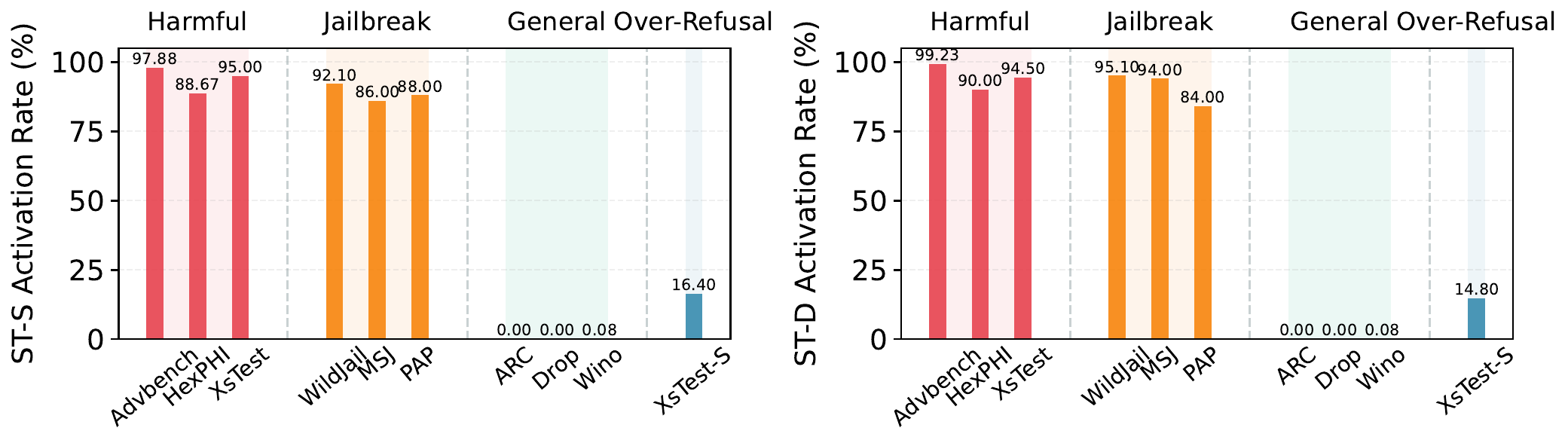}
    \caption{The structured trigger mechanism activation rates of the ST-S and ST-D models across different benchmark categories.}
    \label{fig:activation}
\end{figure}

Figure~\ref{fig:activation} reports the structured trigger mechanism activation rate of ST-S and ST-D across different benchmark categories. The activation rate measures how often the model generates the structured trigger mechanism during inference, reflecting whether the structured trigger mechanism is invoked selectively under risky inputs while remaining silent for benign queries.

    
    
    

Both ST-S and ST-D exhibit high activation probability on risky inputs. On harmful and jailbreak benchmarks, the activation rate remains consistently high and approaches full activation, indicating that the models reliably detect potentially unsafe scenarios and insert the safety analysis before producing the final answer. In contrast, the activation rate on general benchmarks remains close to zero for both models, showing that the structured trigger mechanism is rarely misactivated on benign tasks and thus does not unnecessarily introduce extra safety analysis or interfere with normal user experience. 
The activation rate on the Over-Refusal benchmark is low, but not as close to zero as that of the general category, which is quite reasonable. When the model encounters queries whose risk is uncertain, it proactively invokes the structured trigger mechanism for safety analysis; however, after concluding that the query is benign, it proceeds to answer rather than refusing. This behavior is consistent wi
th the low over-refusal rates reported in Table~\ref{tab:performance} and Table~\ref{tab:performance_comparison}. ST-D shows a slightly higher and more stable activation rate on risky benchmarks compared to ST-S, suggesting that the preference optimization stage further strengthens the reliability of the triggering behavior under adversarial conditions, while preserving the restraint on benign queries.

\subsubsection{Inference Resource}

\begin{figure}[t]
    \centering
    \includegraphics[width=0.7\columnwidth]{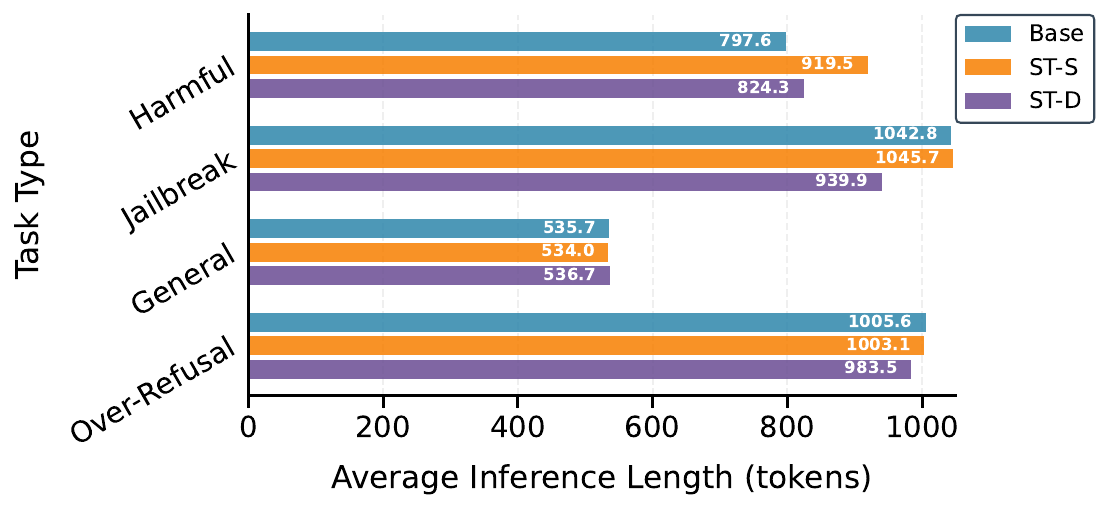}
    \caption{Comparison of inference overhead between the Safe Trigger approach and the base model.}
    \label{fig:inference_resource}
\end{figure}

Figure~\ref{fig:inference_resource} compares the average inference length of Base, ST-S, and ST-D across four task categories. To ensure a fair comparison of inference overhead, the harmful and jailbreak results are computed only on samples whose responses are judged as safe. This control is important because if unsafe outputs were included, the Safe Trigger approach would convert a large portion of responses into safe ones, and safe responses are naturally shorter than unsafe, detailed completions. This evaluation isolates the effect of the structured trigger mechanism (\textit{<safe>}) itself and reveals how much additional inference cost it introduces.

For harmful queries, ST-S generates longer outputs than Base, which is expected because the structured trigger mechanism (\textit{<safe>}) explicitly inserts a structured safety analysis before the final answer. After preference optimization, \textit{<safe>} becomes more concise and targeted, so ST-D is only slightly longer than Base, indicating improved efficiency in safety analysis generation.
For jailbreak queries under the safe-only setting, ST-S shows an inference length that is nearly identical to Base, while ST-D exhibits a clear reduction. Based on our experimental observations, this is because the Base model often gets trapped in attacker-crafted narratives during jailbreak attempts and generates redundant explanatory tokens. In contrast, Safe Trigger emphasizes directly extracting and analyzing the core issues, which significantly eliminates unnecessary output.
For general and over-refusal categories, the response lengths of ST-S and ST-D remain nearly identical to Base, suggesting that Safe Trigger introduces negligible overhead on benign inputs and has minimal impact on overall generation behavior when safety triggering is not required.

\section{Related Work}
A substantial body of work has explored jailbreak attacks~\cite{andriushchenko2024jailbreaking,zou2023universal,huang2023catastrophic}, privacy exposure risks~\cite{liu2024dpzero,miao2025realworld,wang2025deconstruct,luo2025shadow,hu2025membership,bao2025mapeunlearn,hu2024eraser}, and safety alignment~\cite{dai2024safe,wachi2024stepwise,li2024think,phan2025thinktwicegenerateonce,si2025thinkrefusaltriggering} for LLMs. In contrast, there is still less research work focusing on the safety of LRMs~\cite{huang2025safetytaxsafetyalignment,zhou2025hidden,guan2024deliberative,jiang2025safechain,wang2025star,jeung2025safepath}. 

For LRMs' safety,~\cite{huang2025safetytaxsafetyalignment} first introduced the concept of Safety Tex for safety alignment in Large Reasoning Models, highlighting that for reasoning models, safety alignment naturally leads to a reduction in reasoning capabilities.~\cite{zhou2025hidden} systematically evaluate the safety vulnerability of DeepSeek R1 and find that its enhanced reasoning capabilities can inadvertently amplify harmful outputs compared to vanilla LLMs.~\cite{jiang2025safechain} evaluate the safety of LRMs and find that reducing the thinking content of LRMs to zero could effectively enhance their safety. Additionally, they construct a 40k CoT dataset (SafeChain), which contains reasoning responses from the distilled Llama-70B with DeepSeek R1, to improve the safety of LRMs via fine-tuning. In contrast to the large quantity of CoT data in SafeChain,~\cite{wang2025star} enhance the safety of LRMs via fine-tuning LRMs with 1k high-quality and source-diverse CoT data (STAR1) containing the deliberative reasoning content regarding safety policies~\cite{guan2024deliberative}, which leads to better safety improvement than SafeChain.~\cite{guan2024deliberative} similarly integrate SFT and RL by fine-tuning models on a CoT dataset of safety policies and incorporating safety-aware rewards during RL optimization. The construction of the CoT dataset is time-consuming, and their method is inferior to Star1~\cite{wang2025star} compared in this work. 

Despite initial progress, the above safety alignment methods rely on manual or advanced closed source models to carefully construct CoT training data, which can result in expensive resource overhead.
~\cite{jeung2025safepath} fine-tunes the model to emit a short safety prefix at the start of reasoning when faced with harmful prompts, while leaving the remaining reasoning trajectory unconstrained. It is regarded as a current state-of-the-art approach, achieving strong safety gains with minimal impact on reasoning depth and inference overhead.
However, we found that safety considerations in the reasoning process do not always translate effectively into safe final responses, indicating substantial room for improvement in the safety alignment of LRMs.
Our method significantly improves safety alignment by explicitly activating and strengthening the Latent Safety Awareness of LRMs, while maintaining strong general-purpose performance. It operates without manual annotation or reliance on closed-source models, making it both resource-efficient and highly scalable.


\section{Conclusion}
This paper identifies Latent Safety Awareness in LRMs and proposes the Safe Trigger approach to explicitly and efficiently leverage this capability. 
We introduce Safe Trigger SFT to teach LRMs to selectively invoke a structured trigger mechanism under potentially risky inputs, and further propose Safe Trigger DPO to strengthen the guiding effectiveness and stability in shaping the final answer. 
Experiments show that the Safe Trigger approach substantially improves safety while preserving general capability. 
The proposed pipeline is fully self-bootstrapped, yielding a scalable, low-cost, and stable solution for large-scale safety alignment.

\section{Acknowlegments}
This research is supported in part by the “Pioneer” and “Leading Goose” R\&D Program of Zhejiang (Grant No. 2024C01169), the Kunpeng–Ascend Science and Education Innovation Excellence/Incubation Center, the National Natural Science Foundation of China (Grant No. 6244123), the National Natural Science Foundation of China under Grant U2441240 (“Ye Qisun” Science Foundation), the Li Auto Inc., and the NGICS Comprehensive Security Project for Industrial Control Systems.

\bibliographystyle{splncs04}
\bibliography{mybibliography}
%




\end{document}